\pdfoutput=1

\documentclass[11pt]{article}

\usepackage[]{acl}

\usepackage{times}
\usepackage{latexsym}

\usepackage[T1]{fontenc}

\usepackage[utf8]{inputenc}

\usepackage{microtype}

\usepackage{inconsolata}

\usepackage{hyperref}
\usepackage{url}

\usepackage{graphicx}
\usepackage{subfigure}
\usepackage{wrapfig}

\usepackage{amsmath}
\usepackage{amstext}
\usepackage{amsfonts}
\usepackage{bm}

\usepackage{bbm}
\usepackage{multirow}
\usepackage{booktabs}
\usepackage{array}
\usepackage{caption}
\usepackage{multirow}
\usepackage{booktabs}
\usepackage{array}
\usepackage{caption}
\usepackage{color}
\usepackage{colortbl}
\usepackage{tablefootnote}
\usepackage{adjustbox}

\usepackage{caption}
\usepackage{algorithm}
\usepackage{algpseudocode}

\newcommand{\mydarkcolor}[1]{\textcolor[RGB]{64,101,149}{#1}}
\algnewcommand{\LineComment}[1]{\Statex ~~~~~~\textsc{//}~\textit{#1}}

\usepackage{enumitem}
\setenumerate[1]{itemsep=0pt,partopsep=0pt,parsep=\parskip,topsep=5pt}
\setitemize[1]{itemsep=0pt,partopsep=0pt,parsep=\parskip,topsep=5pt}
\setdescription{itemsep=0pt,partopsep=0pt,parsep=\parskip,topsep=5pt}

\usepackage{soul}

\usepackage{tikz}
\usepackage[edges]{forest}
\definecolor{hidden-draw}{RGB}{64,101,149}
\definecolor{hidden-pink}{RGB}{231,239,250}

\usepackage[mathscr]{euscript}
\newcommand{\M}{\mathcal{M}}

\usepackage{amssymb}  
\usepackage{pifont}

\newcommand{\greenyes}{\textcolor{green}{\ding{51}}}
\newcommand{\redno}{\textcolor{red}{\ding{55}}}
\newcommand{\cpm}[1]{\textcolor{gray}{$_{\pm #1}$}}

\usepackage{ulem}

\title{\texorpdfstring{\includegraphics[width=18pt]{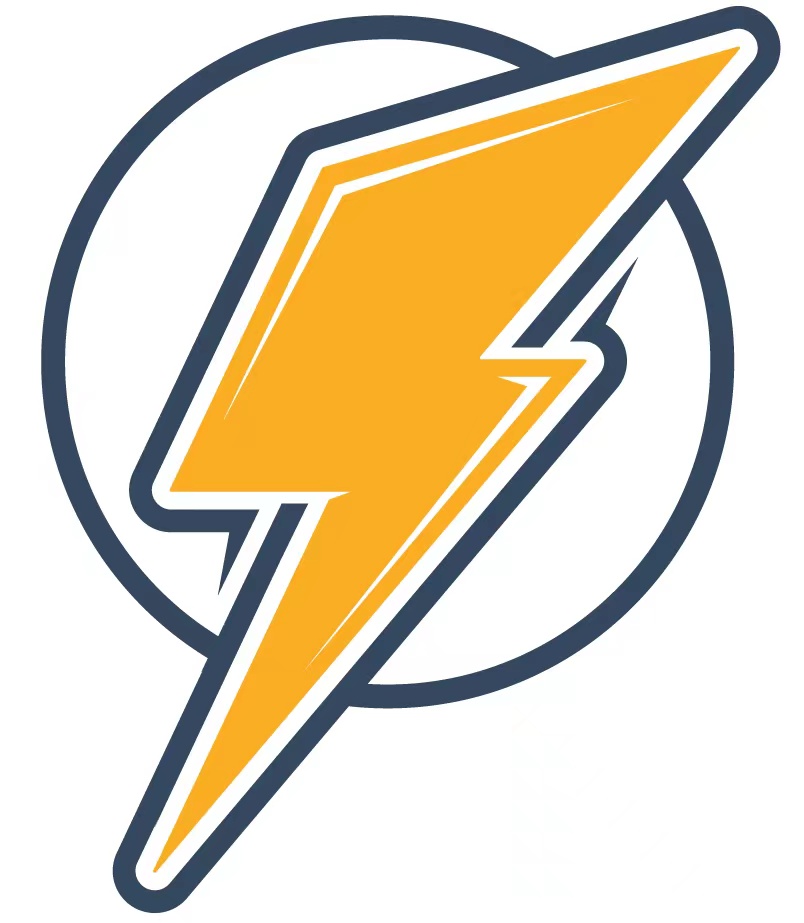}}{}~\textit{Unlocking Efficiency in Large Language Model Inference:}\\A Comprehensive Survey of Speculative Decoding}

\author{{Heming Xia}\textsuperscript{\rm 1}, {Zhe Yang}\textsuperscript{\rm 2}, {Qingxiu Dong}\textsuperscript{\rm 2}, {Peiyi Wang}\textsuperscript{\rm 2},\\
{\textbf{Yongqi Li}}\textsuperscript{\rm 1}, {\textbf{Tao Ge}}\textsuperscript{\rm 3}, {\textbf{Tianyu Liu}}\textsuperscript{\rm 4}, {\textbf{Wenjie Li}}\textsuperscript{\rm 1}, {\textbf{Zhifang Sui}}\textsuperscript{\rm 2}\\
  \textsuperscript{\rm 1}Department of Computing, The Hong Kong Polytechnic University \\
  \textsuperscript{\rm 2}National Key Laboratory for Multimedia Information Processing, Peking University\\
  \textsuperscript{\rm 3}Microsoft Research Asia\quad \textsuperscript{\rm 4}Alibaba Group\\ 
  {\tt \{he-ming.xia\}@connect.polyu.hk; \{yz\_young\}@pku.edu.cn}
}
  
\begin{document}
\maketitle
\begin{abstract}
To mitigate the high inference latency stemming from autoregressive decoding in Large Language Models (LLMs), Speculative Decoding has emerged as a novel decoding paradigm for LLM inference. 
In each decoding step, this method first drafts several future tokens efficiently and then verifies them in parallel. Unlike autoregressive decoding, Speculative Decoding facilitates the simultaneous decoding of multiple tokens per step, thereby accelerating inference. 
This paper presents a comprehensive overview and analysis of this promising decoding paradigm. We begin by providing a formal definition and formulation of Speculative Decoding. Then, we organize in-depth discussions on its key facets, such as drafter selection and verification strategies. Furthermore, we present a comparative analysis of leading methods under third-party testing environments. We aim for this work to serve as a catalyst for further research on Speculative Decoding, ultimately contributing to more efficient LLM inference.\footnote{The relevant papers will be regularly updated at \url{https://github.com/hemingkx/SpeculativeDecodingPapers}.}
\end{abstract}


\section{Introduction}
Large Language Models (LLMs) have achieved remarkable proficiency in a range of downstream tasks~\cite{openai:2023gpt4, Hugo:2023llama, Hugo:2023llama2, vicuna2023, mistral}. They are progressively evolving as the cornerstone of comprehensive API interfaces (e.g., ChatGPT\footnote{\url{https://chat.openai.com}}), offering human life services and guidance through real-time human-machine interactions. However, the inference latency of these sizable models has emerged as a substantial obstacle restricting their broader applications. This latency primarily arises from the token-by-token generation necessitated by autoregressive decoding, resulting in an escalation of the inference latency with both the length of the generated sequence and the model's scale.

\begin{figure}[t]
\centering
    \includegraphics[width=0.95\columnwidth]{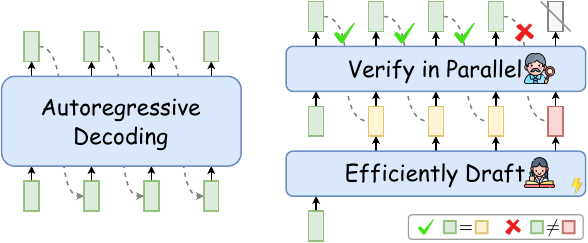}
    \caption{In contrast to autoregressive decoding (\textit{left}) that generates sequentially, Speculative Decoding (\textit{right}) first \textit{efficiently drafts} multiple tokens and then \textit{verifies} them \textit{in parallel} using the target LLM. Drafted tokens after the bifurcation position (\textit{e.g.,} \includegraphics[width=7pt]{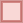}~) will be discarded to guarantee the generation quality.}
    \label{fig:specdec-intro}
\end{figure}
 
To accelerate LLM inference, an innovative inference paradigm, Speculative Decoding has been introduced~\cite{Stern:2018blockwise, xia:2022specdec, Leviathan:2023specdec, Chen:2023specsampling}. As shown in Figure~\ref{fig:specdec-intro}, in each decoding step, Speculative Decoding first efficiently drafts multiple tokens as speculation of future decoding steps of the target LLM and then utilizes the LLM to verify all drafted tokens in parallel. Only those tokens that meet the LLM's verification criterion are accepted as final outputs to guarantee generation quality. 

\begin{figure*}[t]
\centering
\includegraphics[width=0.95\textwidth]{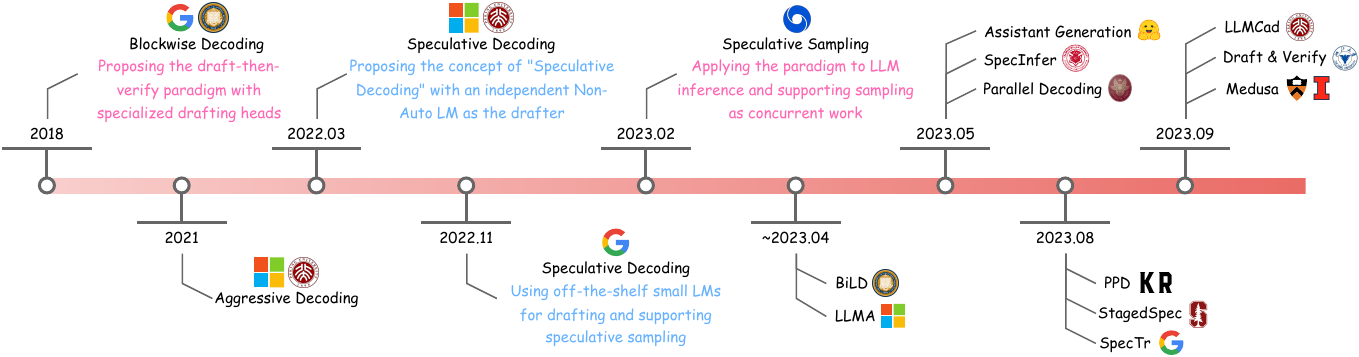}
\caption{Timeline illustrating the evolution of Speculative Decoding. After 2022, Speculative Decoding was formally introduced as a general decoding paradigm to accelerate LLM inference and garnered widespread attention.}
\label{fig:timeline}
\end{figure*}

Speculative Decoding is founded upon two key observations about LLM inference: 1) many easy tokens can be predicted with less computational overhead (e.g., using a smaller model), and 2) LLM inference is highly memory bandwidth bound~\cite{Patterson:2004latencybandwith, Shazeer:2019memorybandwith} with the main latency bottleneck arising from memory reads/writes of LLM parameters rather than arithmetic computations. Drawing on these observations, Speculative Decoding adapts the concept of \textit{speculative execution}\footnote{Speculative execution ~\cite{Burton:1985specexe, John:2012specexe} is an optimization technique used in computer architecture where tasks are performed in advance and subsequently verified for their necessity, thereby circumventing the delays inherent in sequential task execution.} to focus LLMs' efforts on the validation of pre-drafted tokens, substantially diminishing the need for frequent memory operations of LLM parameters, thereby improving inference efficiency. 

While Speculative Decoding shows promise, it raises several critical questions that warrant further investigation. For instance, how to design an optimal drafter to strike a balance between speculation accuracy and drafting efficiency~\cite{xia:2022specdec, Zhou:2023distillspec, Li:2024eagle}. Additionally, it is essential to assess whether the verification criterion can maintain both generation parallelism and output quality~\cite{Miao:2023specinfer, medusa}. Furthermore, since existing methods are evaluated under disparate testing conditions, a unified benchmark is needed to facilitate realistic speedup expectations within the research community.

Amid the rapid expansion of research in Speculative Decoding, this work makes the first attempt to present a survey of this field, aiming to raise awareness among academics about the latest advancements. We provide a systematic categorization of current research and an in-depth analysis of relevant studies. Moreover, we introduce Spec-Bench, a comprehensive benchmark to assess Speculative Decoding methods in diverse application scenarios. Our contributions can be summarized as follows:

\begin{enumerate}[label={(\arabic*)}]
    \item \textbf{\textit{First survey:}} To our knowledge, we are the first to present a comprehensive survey on Speculative Decoding;
    \item \textbf{\textit{Formal definition:}} We furnish a formal definition and formulation of Speculative Decoding, laying the groundwork for future research.
    \item \textbf{\textit{New taxonomy:}} We provide a systematic taxonomy for Speculative Decoding, offering an organized categorization of existing work.
    \item \textbf{\textit{Spec-Bench:}} We introduce Spec-Bench, an extensive benchmark designed for assessing Speculative Decoding, enabling a comparative evaluation of leading methodologies.
\end{enumerate}

We hope that this work can serve as an essential guide for newcomers and motivate future research.
\section{Overview}
This paper offers a comprehensive survey of Speculative Decoding. We commence by introducing the early stages of Speculative Decoding research~(\S\ref{sec:timeline}), illustrated by a timeline of its evolution (as shown in Figure~\ref{fig:timeline}). This is followed by a formal definition and formulation of Speculative Decoding~(\S\ref{sec:definition}). Then, we delve into a detailed discussion of leading techniques, including the selection of draft models~(\S\ref{sec:drafting}), verification strategies~(\S\ref{sec:verification}), and alignment between the drafter and the target LLM~(\S\ref{sec:alignment}). Moreover, we introduce Spec-Bench, an extensive evaluation benchmark designed for assessing the acceleration effect of Speculative Decoding~(\S\ref{sec:comparisons}).

\section{Evolution of Speculative Decoding}
\label{sec:timeline}
This section discusses the motivation behind Speculative Decoding~(\S\ref{sec:motivation}) and then provides a detailed introduction to early attempts in this field~(\S\ref{sec:pioneering_efforts}).

\subsection{Motivation}
\label{sec:motivation}
The widespread adoption of LLMs has established autoregressive decoding as the \textit{de facto} standard to LLM inference~\cite{PaLM:2023, openai:2023gpt4, jiang2024mixtral}. However, autoregressive decoding is limited by its inference latency, which primarily stems from the memory-bound computation of LLMs~\cite{Patterson:2004latencybandwith, Shazeer:2019memorybandwith}. Specifically, the main latency bottleneck of each decoding step is not due to computational operations but arises from the necessity to transfer all LLM parameters from High-Bandwidth Memory (HBM) to the on-chip cache of modern accelerators like GPUs. This process, which generates only one token per step, leads to the underutilization of these accelerators and results in inefficiencies.

\subsection{Pioneering \textit{Draft-then-Verify} Efforts}
\label{sec:pioneering_efforts}
To mitigate the above issue, an intuitive way involves leveraging idle computational resources to enhance parallelism in LLM inference. To this end, \citet{Stern:2018blockwise} introduced Blockwise Decoding, an approach that incorporates extra feedforward neural (FFN) heads atop the Transformer decoder, enabling the simultaneous \textit{drafting} of multiple tokens per step. These tokens are then \textit{verified} by the original LLM \textit{in parallel}, ensuring that the outputs align with those of the original LLM. As a pioneering work proposing the \textit{Draft-then-Verify} paradigm, Blockwise Decoding effectively reduces the number of required LLM calls by increasing generation parallelism, thereby accelerating inference.

To further unleash the potential of this paradigm, \citet{xia:2022specdec} introduced Speculative Decoding (SpecDec), which utilizes an independent drafter, notably a specialized Non-Autoregressive Transformer, to perform the drafting task both accurately and efficiently. Moreover, this method presented an innovative strategy that relaxes the rigid verification criterion, thereby increasing the acceptance rate of drafted tokens. Impressively, SpecDec achieves around 5$\times$ speedup over autoregressive decoding with comparable quality, underscoring the substantial potential of Speculative Decoding.

\begin{algorithm}[t]
\small
\caption{Autoregressive Decoding}
\label{algo:ar}
\begin{algorithmic}[1]
\Require
Language model $\M_q$, input sequence $x_1,\dots,x_t$, and target sequence length $T$;
\State{\textbf{initialize} $n \gets t$}
\While{$n< T$}
\State Set $q_{n+1} \gets \M_q\left(x\mid x_{< n+1}\right)$ 
\State Sample $x_{n+1} \sim q_{n+1}$ 
\State $n \gets n + 1$
\EndWhile
\end{algorithmic}
\end{algorithm}

Following SpecDec, \citet{Leviathan:2023specdec} and \citet{Chen:2023specsampling} made concurrent contributions by proposing Speculative Sampling, expanding this paradigm to encompass the lossless acceleration of various sampling methods. These approaches employed smaller LMs from the same series (e.g., \texttt{T5-small}) to speed up the inference of their larger counterparts (e.g., \texttt{T5-XXL}). Unlike previous work, these \textit{off-the-shelf} small LMs do not require additional training, enabling the rapid adoption of Speculative Decoding in LLM acceleration. This advancement has elevated Speculative Decoding to the forefront of LLM efficiency research, attracting widespread interest within the NLP community.

To sum up, these pioneering efforts in Speculative Decoding have gradually solidified the \textit{Draft-then-Verify} paradigm, showcasing its promising potential in LLM acceleration. We provide a detailed categorization and discussion of these studies and subsequent research in the following sections.

\section{Formulation and Definition}
\label{sec:definition}
In this section, we first provide a concise overview of standard autoregressive decoding~(\S\ref{sec:ar}). Then, we offer an in-depth exposition of Speculative Decoding~(\S\ref{sec:specdec}), which encompasses a formal definition, a comprehensive description of the methodology, and a detailed elaboration of the algorithm.

\begin{algorithm}[t]
\small
\caption{Speculative Decoding}
\label{algo:specdec}
\begin{algorithmic}[1]
\Require
Target language model $\M_q$, draft model $\M_p$, input sequence $x_1, \dots, x_t$, block size $K$, target sequence length $T$, drafting strategy $\textsc{Draft}$, verification criterion $\textsc{Verify}$, and correction strategy $\textsc{Correct}$;
\State{\textbf{initialize} $n \gets t$}
\While{$n< T$}
\mydarkcolor{\LineComment{Drafting: obtain distributions from $\M_p$ efficiently}}
\State Set $p_1, \dots, p_K \gets \textsc{Draft}\left(x_{\leq n}, \M_p\right)$
\mydarkcolor{\LineComment{Drafting: sample $K$ drafted tokens}}
\State Sample $\widetilde{x}_i \sim p_i, i=1, \dots, K$
\mydarkcolor{\LineComment{Verification: compute $K+1$ distributions in parallel}}
\State Set $q_{i} \gets \M_q\left(x\mid x_{\leq n}, \widetilde{x}_{<i}\right), i=1, \dots, K+1$
\mydarkcolor{\LineComment{Verification: verify each drafted token}}
\For{$i=1:K$}
\If{$\textsc{Verify}\left(\widetilde{x}_i, p_{i}, q_{i}\right)$}
\State Set ${x}_{n+i} \gets\widetilde{x}_i$ and $n \gets n + 1$
\Else
\State ${x}_{n+i} \gets \textsc{Correct}\left(p_{i}, q_{i}\right)$
\State and Exit for loop.
\EndIf
\EndFor
\State If all drafted tokens are accepted, sample next token ${x}_{n+1} \sim q_{K+1}$ and set $n \gets n + 1$.
\EndWhile
\end{algorithmic}
\end{algorithm}

\tikzstyle{my-box}=[
    rectangle,
    draw=hidden-draw,
    rounded corners,
    text opacity=1,
    minimum height=1.5em,
    minimum width=5em,
    inner sep=2pt,
    align=center,
    fill opacity=.5,
    line width=0.8pt,
]
\tikzstyle{leaf}=[my-box, minimum height=1.5em,
    fill=hidden-pink!80, text=black, align=left,font=\normalsize,
    inner xsep=2pt,
    inner ysep=4pt,
    line width=0.8pt,
]
\begin{figure*}[t!]
    \centering
    \resizebox{\textwidth}{!}{
        \begin{forest}
            forked edges,
            for tree={
                grow=east,
                reversed=true,
                anchor=base west,
                parent anchor=east,
                child anchor=west,
                base=center,
                font=\large,
                rectangle,
                draw=hidden-draw,
                rounded corners,
                align=left,
                text centered,
                minimum width=4em,
                edge+={darkgray, line width=1pt},
                s sep=3pt,
                inner xsep=2pt,
                inner ysep=3pt,
                line width=0.8pt,
                ver/.style={rotate=90, child anchor=north, parent anchor=south, anchor=center},
            },
            where level=1{text width=7em,font=\normalsize,}{},
            where level=2{text width=8em,font=\normalsize}{},
            where level=3{text width=6em,font=\normalsize,}{},
            where level=4{text width=6em,font=\normalsize,}{},
            [
                Speculative Decoding, ver
                [
                    Drafting (\S\ref{sec:drafting}), ver
                    [
                        Independent\\ Drafting (\S\ref{sec:independent_drafting})
                        [
                            Fine-tuned\\ Drafter
                            [
                                SpecDec~\cite{xia:2022specdec}{, }BiLD~\cite{Kim:2023bild}{, }SpecInfer~\cite{Miao:2023specinfer}{, }\\Online Speculative~\cite{Liu:2023onlinespec}{, }DistillSpec~\cite{Zhou:2023distillspec}
                                , leaf, text width=35em
                            ]
                        ]
                        [
                            Tuning-free\\ Drafter
                            [
                                Speculative Decoding~\cite{Leviathan:2023specdec}{, }StagedSpec~\cite{Spector:2023stagedspec}{, }\\SpS~\cite{Chen:2023specsampling}{, }SpecTr~\cite{Sun:2023spectr}{, }REST~\cite{He:2023REST}{, }\\CS. Drafting~\cite{chen2023cascadespec}{, }MCSD~\cite{Yang:2024mcsd}
                                , leaf, text width=35em
                            ]
                        ]
                    ]
                    [
                        Self-Drafting (\S\ref{sec:self_drafting})
                            [
                                FFN Heads
                                [
                                    Blockwise~\cite{Stern:2018blockwise}{, }Medusa~\cite{medusa}{, }EAGLE~\cite{Li:2024eagle}
                                    , leaf, text width=35em
                                ]
                            ]
                            [
                                Early Exiting
                                [
                                    PPD~\cite{Yang:2023PPD}{, }Self-Speculative~\cite{Zhang:2023draftverify}{, }\\SPEED~\cite{Hooper:2023speed}
                                    , leaf, text width=35em
                                ]
                            ]
                            [
                                Mask-Predict
                                [
                                    Parallel Decoding~\cite{Santilli:2023paralleldecoding}{, }Lookahead Decoding~\cite{Fu:2023lookahead}{, }\\PaSS~\cite{Monea:2023PaSS}
                                    , leaf, text width=35em
                                ]
                            ]
                    ]
                ]
                [
                    Verification (\S\ref{sec:verification}), ver
                        [
                            Greedy\\Decoding (\S\ref{sec:greedy_decoding})
                            [
                                Lossless
                                [
                                    Blockwise~\cite{Stern:2018blockwise}{, }SpecDec~\cite{xia:2022specdec}{, }Parallel Decoding\\~\cite{Santilli:2023paralleldecoding}{, }PPD~\cite{Yang:2023PPD}{, }SPEED~\cite{Hooper:2023speed}{, }\\Self-Speculative~\cite{Zhang:2023draftverify}{, }Lookahead Decoding~\cite{Fu:2023lookahead}
                                    , leaf, text width=35em
                                ]
                            ]
                            [
                                Approximate
                                [
                                    Blockwise~\cite{Stern:2018blockwise}{, }SpecDec~\cite{xia:2022specdec}{, }BiLD~\cite{Kim:2023bild}
                                    , leaf, text width=35em
                                ]
                            ]
                        ]
                        [
                            Speculative\\Sampling (\S\ref{sec:nucleus_sampling})
                            [
                                Lossless
                                [
                                    Speculative Decoding~\cite{Leviathan:2023specdec}{, }DistillSpec~\cite{Zhou:2023distillspec}{, }\\Online Speculative~\cite{Liu:2023onlinespec}{, }SpS~\cite{Chen:2023specsampling}{, }CS. Drafting\\\cite{chen2023cascadespec}{, }PaSS~\cite{Monea:2023PaSS}{, }MCSD~\cite{Yang:2024mcsd}
                                    , leaf, text width=35em
                                ]
                            ]
                            [
                                Approximate
                                [
                                    Speculative Decoding~\cite{Leviathan:2023specdec}{, }DistillSpec~\cite{Zhou:2023distillspec}
                                    , leaf, text width=35em
                                ]
                            ]
                        ]
                        [   
                            Token Tree \\ Verification (\S\ref{sec:token_tree_verification})
                            [
                                SpecInfer~\cite{Miao:2023specinfer}{, }StagedSpec~\cite{Spector:2023stagedspec}{, }SpecTr~\cite{Sun:2023spectr}{, }\\REST~\cite{He:2023REST}{, }Medusa~\cite{medusa}{, }EAGLE~\cite{Li:2024eagle}
                                , leaf, text width=42.7em
                            ]
                        ]
                ]
            ]
        \end{forest}
    }
    \caption{Taxonomy of Speculative Decoding.}
    \label{fig:taxo_of_specdec}
\end{figure*}
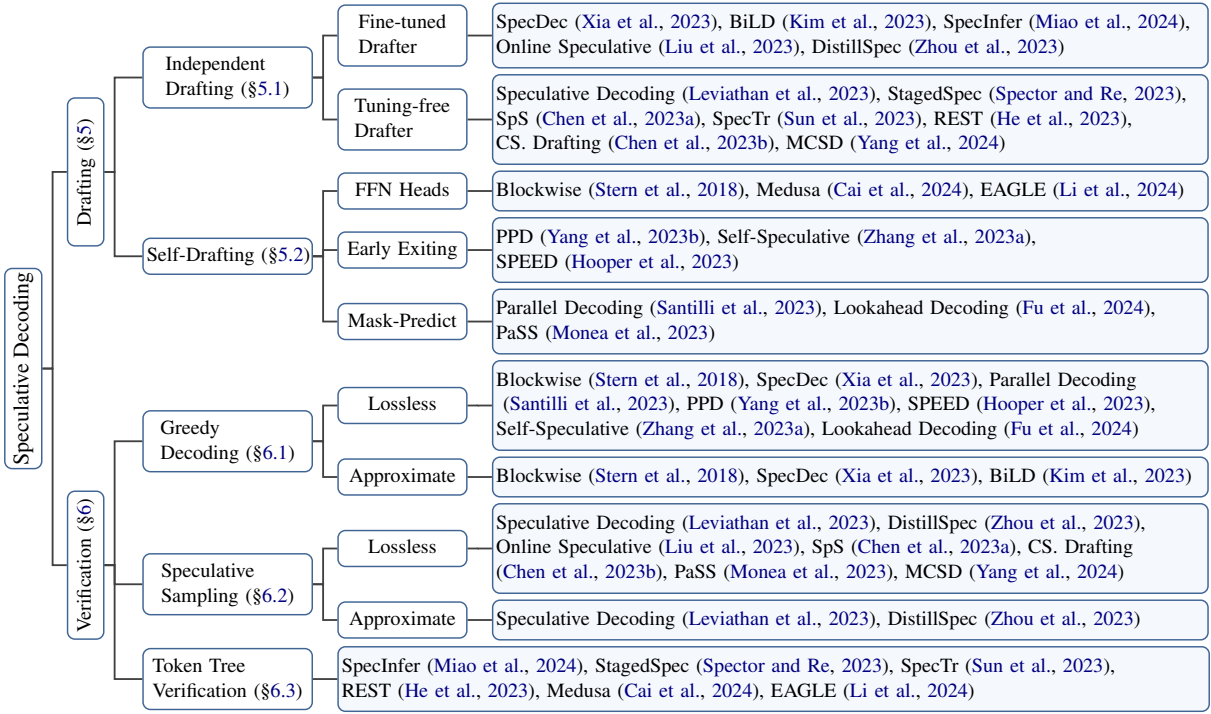

\subsection{Autoregressive Decoding}
\label{sec:ar}
Transformer-based LLMs typically make generations in an autoregressive manner. Given an input sequence $x_1, \dots, x_t$, an autoregressive language model $\M_{q}$ generates the next token according to:
\begin{equation}\label{eq:ar}
    x_{t+1} \sim q_{t+1} = \M_q\left(x\mid x_{<t+1}\right),
\end{equation}
where $q$ is the conditional probability distribution calculated by $\M_{q}$ and $x_{t+1}$ denotes the next token sampled from $q_{t+1}$. We illustrate a detailed process in Algorithm~\ref{algo:ar}.

As discussed in Section~\ref{sec:timeline}, while the standard autoregressive decoding offers desirable generation quality, it is bounded by memory bandwidth, resulting in low utilization of modern accelerators. In this process, each memory-bound LLM call (i.e., an LLM forward step) produces merely a single token for the entire sequence, making the whole generation inefficient and time-consuming.

\subsection{Speculative Decoding}
\label{sec:specdec}
Following \citet{xia:2022specdec}, \citet{Leviathan:2023specdec}, and \citet{Chen:2023specsampling}, we here provide a formal definition of Speculative Decoding: 

\begin{quote}
    Speculative Decoding is a \textit{Draft-then-Verify} decoding paradigm in which, at each decoding step, it first \textit{efficiently drafts} multiple future tokens and then \textit{verifies} all these tokens \textit{in parallel} using the target LLM to speed up inference.
\end{quote}

We formulate a detailed Speculative Decoding process in Algorithm~\ref{algo:specdec}. Subsequently, we delve into the two fundamental substeps integral to this paradigm -- \textit{drafting} and \textit{verification}:

\paragraph{Drafting} At each decoding step, Speculative Decoding first efficiently drafts multiple future tokens, as a speculation of the target LLM's output. Formally, given an input sequence $x_1, \dots, x_t$ and the target LLM $\M_{q}$, this paradigm employs an efficient draft model $\M_{p}$ (e.g., a smaller LM) to decode the next $K$ drafted tokens:
\begin{equation}
\begin{aligned}
p_1, \dots, p_K &= \textsc{Draft}\left(x_{\leq t}, \mathcal{M}_p\right), \\
\widetilde{x}_i &\sim p_i, \quad i=1, \dots, K,
\end{aligned}
\label{eq:draft-computing}
\end{equation}
where $\textsc{Draft}(\cdot)$ denotes various drafting strategies that we will discuss in Section~\ref{sec:drafting}, $p$ is the conditional probability distribution calculated by $\M_{p}$, and $\widetilde{x}_i$ denotes the drafted token sampled from $p_i$.

\paragraph{Verification} Subsequently, these drafted tokens are verified by the target LLM $\M_{q}$ in parallel. Formally, given the input sequence $x_1, \dots, x_t$ and the draft $\widetilde{x}_1, \dots, \widetilde{x}_K$, Speculative Decoding utilizes $\M_{q}$ to compute $K+1$ probability distributions simultaneously:
\begin{align}\label{eq:verify_computing}
q_{i} = \M_q\left(x\mid x_{\leq t}, \widetilde{x}_{<i}\right), i=1, \dots, K+1.
\end{align}

\begin{table*}[t]
\centering
\small
\begin{tabular}{@{}>{\centering\arraybackslash}m{55pt}>{\centering\arraybackslash}m{110pt}>{\arraybackslash}m{263pt}@{}}
\toprule
\bf Methods &\bf $\textsc{Draft}\left(x_{\leq t}, \mathcal{M}_p\right)$ & \bf Drafter Type\\\midrule
Parallel Drafting & $p_1, \dots, p_K = \M_p\left(x\mid x_{\leq t}\right)$ &FFN Heads~\cite{Stern:2018blockwise, medusa}, Non-Autoregressive LM~\cite{xia:2022specdec}, Mask-Predict~\cite{Santilli:2023paralleldecoding, Fu:2023lookahead}\\\midrule
Autoregressive Drafting & $p_{i} = \M_p\left(x\mid x_{\leq t}, \widetilde{x}_{<i}\right), i=1, \dots, K$ & Small LMs~\cite{Leviathan:2023specdec, Chen:2023specsampling}, Early Exiting~\cite{Yang:2023PPD}, Layer Skipping~\cite{Zhang:2023draftverify}\\
\bottomrule
\end{tabular}
\caption{Summary of formulations for various drafting strategies in Speculative Decoding. We categorize these methods into two distinct groups based on their formulations: \textit{parallel drafting} and \textit{autoregressive drafting}.}
\label{tab:drafting}
\end{table*}

Then, each drafted token $\widetilde{x}_i$ is verified by a specific criterion $\textsc{Verify}\left(\widetilde{x}_i, p_{i}, q_{i}\right)$. Only those tokens that meet the criterion are selected as final outputs, ensuring quality consistent with the target LLM's standards. Otherwise, the first drafted token $\widetilde{x}_c$ that fails the verification will be corrected by the strategy $\textsc{Correct}\left(p_{c}, q_{c}\right)$. All drafted tokens after position $c$ will be discarded, to guarantee the high quality of the final outputs. If all tokens pass verification, an additional token $x_{t+K+1}$ will be sampled from $q_{K+1}$ as Eq.~(\ref{eq:ar}). 

The drafting and verification substeps will be iterated until the termination condition is met, i.e., the \texttt{[EOS]} token is decoded or the sentence reaches the maximal length. 

Notably, the acceleration effect of Speculative Decoding primarily hinges on \textit{the acceptance rate} of drafted tokens at each step. This rate is influenced by several factors, including the draft quality, verification criteria, and the behavior alignment between the drafter and the target LLM. Additionally, the intrinsic efficiency of the drafter itself also contributes to the overall speedup. In subsequent sections, we will delve into these pivotal components of Speculative Decoding, as depicted in Figure~\ref{fig:taxo_of_specdec}, to systematically categorize recent research trends within this promising paradigm.

\section{\texorpdfstring{\includegraphics[width=16pt,height=16pt]{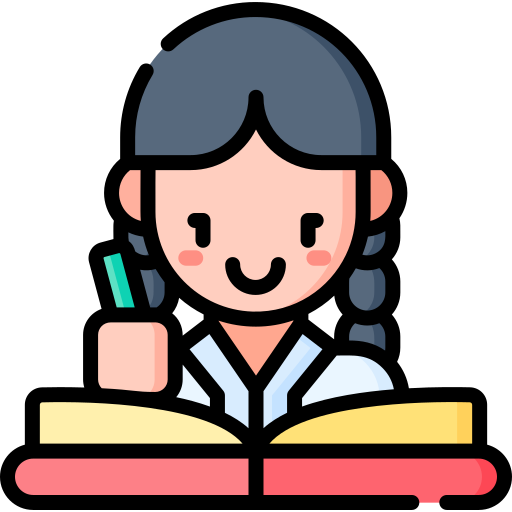}}{}~Drafting}
\label{sec:drafting}
As a vital component of Speculative Decoding, the drafting process has a crucial impact on the speedup of the paradigm. The impact is determined by two key factors: the speculation accuracy of the drafter $\M_{p}$, measured by the average number of accepted tokens per step, and the drafting latency~\cite{Stern:2018blockwise, xia:2022specdec}. How to trade off high speculation accuracy and low drafting latency presents a major challenge in this process. In this section, we classify various drafting strategies into two categories: independent drafting~(\S\ref{sec:independent_drafting}) and self-drafting~(\S\ref{sec:self_drafting}), and summarize their formulations $\textsc{Draft}\left(x_{\leq t}, \mathcal{M}_p\right)$ in Table~\ref{tab:drafting}.

\subsection{Independent Drafting}
\label{sec:independent_drafting}
To strike a balance between speculation accuracy and efficiency, SpecDec~\cite{xia:2022specdec} first proposed utilizing an independent model for drafting. Specifically, it employed a specialized Non-Autoregressive Transformer that drafts multiple tokens simultaneously per step. This model has a deep-shallow encoder-decoder architecture to run efficiently. Despite its strengths, SpecDec requires training a draft model from scratch, which demands an increased computational budget.

Considering the available models in existing LLM series (e.g., OPT~\cite{opt} and LLaMA~\cite{Hugo:2023llama, Hugo:2023llama2}), a more straightforward and efficient approach is directly employing a small LM from the same series as the drafter to accelerate the inference of its larger counterparts~\cite{Leviathan:2023specdec, Chen:2023specsampling, Spector:2023stagedspec, Sun:2023spectr, chen2023cascadespec}. For instance, \citet{Leviathan:2023specdec} utilized \texttt{T5-small} as the drafter, to accelerate the inference of \texttt{T5-XXL}. These \textit{off-the-shelf} small LMs do not require additional training or any modification on model architectures, facilitating the quick adoption of Speculative Decoding. Moreover, since models in the same series share tokenizers, pretraining corpora, and similar training processes, they inherently have an alignment in prediction behaviors.

\begin{table*}[t]
\centering
\small
\resizebox{\linewidth}{!}{%
\begin{tabular}{@{}>{\centering\arraybackslash}m{42pt}>{\centering\arraybackslash}m{128pt}>{\centering\arraybackslash}m{128pt}>{\centering\arraybackslash}m{132pt}@{}}
\toprule
\bf Methods & \bf $\textsc{Verify}\left(\widetilde{x}_i, p_i, q_i\right)$ & \bf $\textsc{Correct}\left(p_c, q_c\right)$ & \bf Representative Work\\\midrule
Greedy Decoding & $\widetilde{x}_i = \arg \max q_i$ & ${x}_{t+c} \gets \arg \max q_c$ & Blockwise Decoding~\cite{Stern:2018blockwise}, SpecDec~\cite{xia:2022specdec}\\
\midrule
Speculative Sampling & $r < \min\left(1, \frac{q_i(\widetilde{x}_i)}{p_i(\widetilde{x}_i)}\right), r \sim U\left[0,1\right]$ & $x_{t+c} \sim \operatorname{norm}(\max\left(0, q_c - p_c\right))$ & Speculative Decoding~\cite{Leviathan:2023specdec}, SpS~\cite{Chen:2023specsampling}\\
\bottomrule
\end{tabular}%
}
\caption{Summary of formulations for various verification strategies in Speculative Decoding.}
\label{tab:verification}
\end{table*}

\subsection{Self-Drafting}
\label{sec:self_drafting}
While leveraging an external draft model offers considerable advantages, this approach necessitates extra effort to either train or identify a draft model that closely aligns with the target LLM. This challenge is intensified when no smaller counterparts of the LLM are available, e.g., LLaMA-7B~\cite{Hugo:2023llama, Hugo:2023llama2}. Furthermore, integrating two distinct models within a single system introduces additional computational complexity, particularly in distributed settings~\cite{medusa}. 

To address the above issues, numerous studies have suggested leveraging the target LLM itself for efficient drafting~\cite{Stern:2018blockwise, Santilli:2023paralleldecoding, Hooper:2023speed, medusa, Fu:2023lookahead, Du:2024glide}. Particularly, Blockwise Decoding~\cite{Stern:2018blockwise} and Medusa~\cite{medusa} incorporated FFN heads atop the Transformer decoder, enabling the parallel token generation per step. Compared with external drafters, these lightweight heads reduce extra computational overhead and are friendly to distributed inference. Another line of research has explored the potential of \textit{early exiting} and \textit{layer skipping} within the target LLM for drafting~\cite{Yang:2023PPD, Zhang:2023draftverify, Hooper:2023speed}. For instance, \citet{Yang:2023PPD} introduced additional subprocesses that exit early during the current decoding step, thereby initiating the drafting of future tokens in advance. Similarly, Self-Speculative~\cite{Zhang:2023draftverify} proposed to adaptively skip several intermediate layers during inference to draft efficiently. 

In contrast to prior work that focused on extending model architectures or altering the inference process, \citet{Santilli:2023paralleldecoding} introduced a simple drafting strategy that directly appends multiple \texttt{[PAD]} tokens to the end of the input prompt to enable parallel generation. However, this approach deviates from LLMs' autoregressive pretraining pattern, leading to suboptimal drafting quality. To tackle this, \citet{Fu:2023lookahead} proposed transforming low-quality drafts into multiple n-grams to improve the speculation accuracy; \citet{Monea:2023PaSS} introduced multiple learnable \texttt{[LA]} tokens and finetuned these token embeddings on a small training dataset to enhance the parallel decoding performance.

\section{\texorpdfstring{\includegraphics[width=16pt,height=16pt]{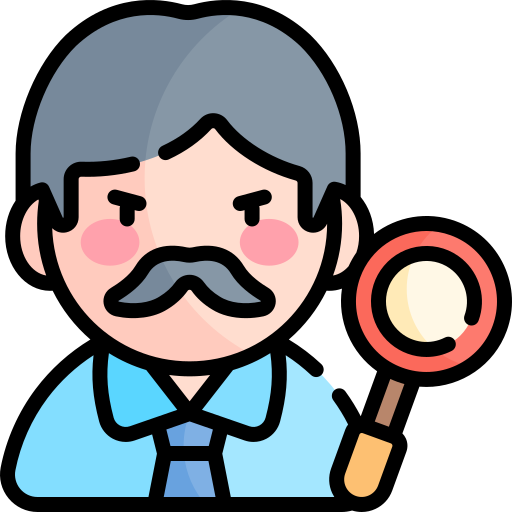}}{}~Verification}
\label{sec:verification}
In each decoding step, the drafted tokens are \textit{verified in parallel} to ensure the outputs align with the target LLM. This process also determines the number of tokens accepted per step, a vital factor impacting the speedup. This section summarizes various verification criteria $\textsc{Verify}\left(\widetilde{x}_i, p_i, q_i\right)$ (as shown in Table~\ref{tab:verification}), encompassing those supporting greedy decoding~(\S\ref{sec:greedy_decoding}) and speculative sampling~(\S\ref{sec:nucleus_sampling}) in LLM inference. Besides, we introduce token tree verification~(\S\ref{sec:token_tree_verification}), an effective strategy to increase the token acceptance rate.

\begin{table*}[t]
\centering
\small
\setlength{\tabcolsep}{3pt}
\resizebox{\linewidth}{!}{
\begin{tabular}{@{}ll|ccccccccc@{}}
\toprule
\multicolumn{2}{l|}{\multirow{2}{*}{\bf Methods}} & \multicolumn{3}{c}{Drafting} & \multicolumn{3}{c}{Verification} &\multirow{2}{*}{\begin{tabular}[c]{@{}c@{}}Target LLM\end{tabular}}&\multirow{2}{*}{\begin{tabular}[c]{@{}c@{}}Speedup\\(reported)\end{tabular}}\\ \cmidrule(lr){3-5} \cmidrule(lr){6-8}
\multicolumn{2}{l|}{} & Approach & Alignment &Tuning-free &Greedy &Sampling &Token Tree\\ \midrule
\multirow{7}{*}{\rotatebox{90}{\textit{Independent-D}}} 
&SpecDec~\cite{xia:2022specdec} &Non-Auto LM &Seq-KD  &\redno &\greenyes   &\redno &\redno &Transformer-base (65M) &$3.9\times \sim 5.1\times$\\
&SpS~\cite{Chen:2023specsampling} &Small LM &-  &\greenyes  &\greenyes&\greenyes &\redno &Chinchilla (70B) &$1.9\times \sim 2.5\times$ \\
&SpecInfer~\cite{Miao:2023specinfer} &Boost-tuned LMs  &Col-BT  &\redno &\greenyes  &\greenyes &\greenyes &LLaMA (30B-65B) &$2.0\times \sim 2.4\times$\\
&DistillSpec~\cite{Zhou:2023distillspec} &Small LM  &KD &\redno  &\greenyes &\greenyes &\redno  &T5-XL (3B) &-\\
&Online Speculative~\cite{Liu:2023onlinespec} &Small LM  &Online-KD &\redno  &\greenyes &\greenyes &\redno  &Vicuna (7B) &-\\
&CS. Drafting~\cite{chen2023cascadespec} &Cascaded LMs  &-  &\greenyes  &\greenyes &\greenyes &\redno &FLAN-T5-xxl (11B) &-\\
&REST~\cite{He:2023REST} &Context Retrieval  &-  &\greenyes  &\greenyes &\greenyes &\greenyes &Vicuna (7B-13B) &$1.6\times \sim 1.8\times$\\\midrule
\multirow{7}{*}{\rotatebox{90}{\textit{Self-D}}} 
&Blockwise Decoding~\cite{Stern:2018blockwise} &FFN Heads &Seq-KD &\redno &\greenyes &\redno &\redno &Transformer-big (213M) &$1.7\times \sim 3.0\times$\\
&Medusa~\cite{medusa} &FFN Heads &Seq-KD &\redno &\greenyes &\greenyes &\greenyes &Vicuna (7B-13B) &$2.2\times \sim 2.3\times$\\\
&PPD~\cite{Yang:2023PPD} &Early Exiting  &-  &\redno  &\greenyes &\redno &\redno  &Vicuna (13B) &$1.1\times \sim 1.5\times$\\
&Self-Speculative~\cite{Zhang:2023draftverify} &Layer Skipping  &-  &\greenyes  &\greenyes &\greenyes &\redno  &LLaMA-2 (13B-70B) &$1.4\times \sim 1.7\times$\\
&Parallel Decoding~\cite{Santilli:2023paralleldecoding} &Mask-Predict  &-  &\greenyes  &\greenyes &\redno &\redno &MBart50 (610M) &$1.0\times \sim 1.1\times$ \\
&Lookahead Decoding~\cite{Fu:2023lookahead} &Mask-P \& N-grams  &-  &\greenyes &\greenyes &\redno &\redno &LLaMA-2 (7B-70B) &$1.5\times \sim 2.3\times$\\
&EAGLE~\cite{Li:2024eagle} &Auto-regression Head  &KD  &\redno &\greenyes &\greenyes &\greenyes &Vicuna (7B-33B) &$2.9\times \sim 3.1\times$\\
\bottomrule
\end{tabular}}
\caption{Summary of Speculative Decoding methods. ``\textit{Independent-D}'' and ``\textit{Self-D}'' denote independent drafting and self-drafting, respectively. ``\textit{Greedy}'', ``\textit{Sampling}'', and ``\textit{Token Tree}'' denote whether the method supports greedy decoding, speculative sampling, and token tree verification, respectively. We list the most representative target LLMs for each method and the speedups in the original paper (if reported), which is obtained with a batch size of 1.}
\label{tab:comparison}
\end{table*}

\subsection{Greedy Decoding}
\label{sec:greedy_decoding}
Early attempts at Speculative Decoding focused on the verification criterion supporting greedy decoding, which guarantees that the outputs are exactly the same as the greedy decoding results of the target LLM~\cite{Stern:2019, Sun:2020SAD, xia:2022specdec}. Formally, given the input sequence $x_1, \dots, x_t$, the drafted tokens $\widetilde{x}_1, \dots, \widetilde{x}_K$, and the computed probability distributions $p_1, \dots, p_{K}$, $q_1, \dots, q_{K}$ as obtained from Eq.~(\ref{eq:draft-computing}) and (\ref{eq:verify_computing}), respectively, the verification criterion on the $i_{th}$ drafted token is formulated as
\begin{equation}\label{eq:verify-greedy}
\widetilde{x}_i = \arg \max q_i,
\end{equation}
where $i=1,\dots,K$. The first position $c$ that the drafted token $\widetilde{x}_c$ fails the verification denotes the \textit{bifurcation} position. The output token at this position ${x}_{t+c}$ will be adjusted by the correction strategy, which simply replaces the drafted token with the LLM's top-1 prediction:
\begin{equation}
{x}_{t+c} \gets \arg \max q_c.
\end{equation}

The verification criterion of greedy decoding is straightforward and clear. Thus, multiple subsequent studies have adopted this criterion to demonstrate the efficacy of their methodologies~\cite{Santilli:2023paralleldecoding, Yang:2023PPD, Hooper:2023speed, Zhang:2023draftverify, Fu:2023lookahead}. However, the strict matching requirement of this criterion often results in the rejection of high-quality drafted tokens, simply because they differ from the top-1 predictions of the target LLM, thereby constraining the speedup of the paradigm.

To tackle this problem, multiple studies have proposed various approximate verification criteria~\cite{Stern:2018blockwise, xia:2022specdec, Kim:2023bild}. Compared with the lossless criterion, these methods slightly relax the matching requirement to trust the drafts more, leading to higher acceptance of drafted tokens. For instance, SpecDec~\cite{xia:2022specdec} only requires the drafted tokens to fall in top-k candidates of the target LLM; BiLD~\cite{Kim:2023bild} proposed a rollback criterion that only rejects drafted tokens when the number of consecutive mismatch tokens exceeds a fixed threshold.

\subsection{Speculative Sampling}
\label{sec:nucleus_sampling}
Following \citet{Stern:2019}, subsequent work extended Speculative Decoding to support various sampling methods~\cite{Leviathan:2023specdec, Chen:2023specsampling}, accelerating the target LLM's inference without changing its output distribution. Formally, given the initial sequence $x_1, \dots, x_t$, the drafted tokens $\widetilde{x}_1, \dots, \widetilde{x}_K$ and the computed distributions $p_1, \dots, p_{K}$, $q_1, \dots, q_{K}$, the verification criterion on the $i_{th}$ drafted token is
\begin{equation}
\begin{aligned}\label{eq:verify-sampling}
r < \min\left(1, \frac{q_i(\widetilde{x}_i)}{p_i(\widetilde{x}_i)}\right), r \sim U\left[0,1\right],
\end{aligned}
\end{equation}
where $r$ denotes a random number drawn from a uniform distribution $U\left[0,1\right]$; $q_i(\widetilde{x}_i)$ and $p_i(\widetilde{x}_i)$ are the probability of $\widetilde{x}_i$ according to $\M_q$ and $\M_p$, respectively; and $i=1, \dots, K$. In other words, this criterion accepts the token $\widetilde{x}_i$ if $q_i(\widetilde{x}_i) \geq p_i(\widetilde{x}_i)$, and in case $q_i(\widetilde{x}_i) < p_i(\widetilde{x}_i)$ it rejects the token with probability $1-\frac{q_i(\widetilde{x}_i)}{p_i(\widetilde{x}_i)}$. The correction strategy resamples the output token at the bifurcation position $c$ from an adjusted distribution:
\begin{equation}
x_{t+c} \sim \operatorname{norm}(\max\left(0, q_c - p_c\right)).
\end{equation}

\citet{Leviathan:2023specdec} and \citet{Chen:2023specsampling} have theoretically proved that this criterion maintains identical output distributions to the target LLM. Thus, it has been widely adopted in subsequent research~\cite{Liu:2023onlinespec, Zhou:2023distillspec, Monea:2023PaSS, chen2023cascadespec}. In addition to the strict requirement, some work has also explored approximate strategies to improve the token acceptance rate~\cite{Leviathan:2023specdec, Zhou:2023distillspec}. For instance, \citet{Leviathan:2023specdec} proposed multiplying $p_i(\widetilde{x}_i)$ in Eq.~(\ref{eq:verify-sampling}) by a lenience parameter $l \in [0,1]$ to slightly relax the criterion.

\subsection{Token Tree Verification}
\label{sec:token_tree_verification}
\begin{figure}[t]
\centering
    \includegraphics[width=0.95\columnwidth]{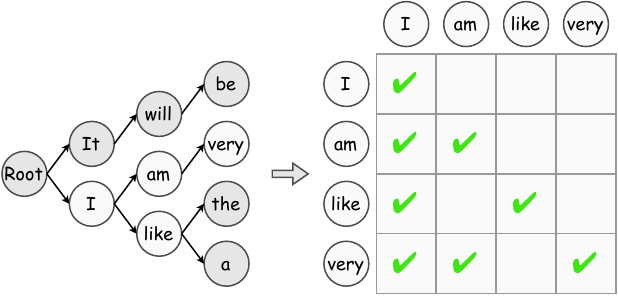}
    \caption{Illustration of the token tree sequences (\textit{left}) and tree attention mask (\textit{right}). For simplicity, we only visualize the attention mask of tokens in white colors.}
    \label{fig:token_tree}
\end{figure}

Contrary to prior verification strategies that focused on a single draft sequence, SpecInfer~\cite{Miao:2023specinfer} proposed \textit{token tree verification}, an effective strategy enabling the target LLM to verify multiple draft sequences in parallel. As illustrated in Figure~\ref{fig:token_tree}, this method first merges multiple candidate draft sequences into a \textit{token tree} by sharing prefixes. It then utilizes a specially designed \textit{tree attention mask} to facilitate the LLM verifying the whole structure in parallel. Recent research has explored various approaches to obtain these candidate draft sequences~\cite{Miao:2023specinfer, medusa, He:2023REST, Li:2024eagle}. For instance, \citet{Miao:2023specinfer} generated diverse draft sequences from different boost-tuned LMs; \citet{medusa} considered the top-k predictions from each FFN head to obtain multiple candidate sequences.

\section{Alignment}
\label{sec:alignment}
As illustrated in Section~\ref{sec:drafting}, the speedup of Speculative Decoding primarily depends on the speculation accuracy, which in turn is influenced by the behavior similarity between the drafter and the target LLM. To enhance this, existing research has explored various knowledge distillation (KD) strategies to align the drafter's outputs with those of the target LLM~\cite{Stern:2018blockwise, xia:2022specdec, Miao:2023specinfer, Liu:2023onlinespec, Kim:2023bild, Zhou:2023distillspec}. Particularly, Blockwise Decoding adopted sequence-level knowledge distillation (Seq-KD)~\cite{kim2016seqkd} for alignment, which trained the drafter on the sentences generated by the target LLM. \citet{Miao:2023specinfer} proposed a collective boost-tuning (Col-BT) strategy, applying Seq-KD to finetune multiple small LMs on the training data and utilizing their aggregated output as drafts to improve the speculation accuracy.

Although Seq-KD is effective, it ignores the probability distributions of the target LLM, leading to performance degradation with sampling methods. To rectify this, recent studies have explored other KD strategies for Speculative Decoding~\cite{Zhou:2023distillspec, Liu:2023onlinespec}. Notably, DistillSpec~\cite{Zhou:2023distillspec} conducted a comprehensive comparison of different KD strategies on Speculative Decoding across various downstream tasks. \citet{Liu:2023onlinespec} proposed an online KD strategy that dynamically aligns the drafter with the target LLM on the fly using the query data.

We summarize the main features of existing Speculative Decoding methods in Table~\ref{tab:comparison}, including the drafter type or the drafting strategy, the alignment approach, supported verification strategies, and the reported speedup, etc.

\section{Spec-Bench}
\label{sec:comparisons}

With the rapid research progress in Speculative Decoding, there is an increasing demand for comparative analysis of leading methods. However, existing approaches are tested using disparate benchmarks, devices, and environments, making fair comparisons impractical. To address this gap, we introduce Spec-Bench -- a comprehensive benchmark for Speculative Decoding covering diverse application scenarios. Based on Spec-Bench, we present a systematic comparison of open-source approaches under third-party testing conditions. Experiments were executed on \textit{the same device and testing environment} to ensure a fair comparison.

\subsection{Benchmark Construction}
To assess Speculative Decoding methods across various scenarios, Spec-Bench encompasses six distinct subtasks: multi-turn conversation, translation, summarization, question answering, mathematical reasoning, and retrieval-augmented generation. 
We composed Spec-Bench by randomly selecting 80 instances from each of six widely used datasets, including MT-bench~\cite{zheng2023mtbench}, WMT14 DE-EN, CNN/Daily Mail~\cite{Nallapati:2016cnndm}, Natural Questions~\cite{kwiatkowski2019:naturalquestions}, GSM8K~\cite{Cobbe:2021gsm8k}, and DPR~\cite{Karpukhin:2020dpr}.
For details on Spec-Bench and the specific experimental setup, please refer to Appendix~\ref{appendix:experimental_details}.

\begin{figure}[t]
    \centering
    \includegraphics[width=0.9\columnwidth]{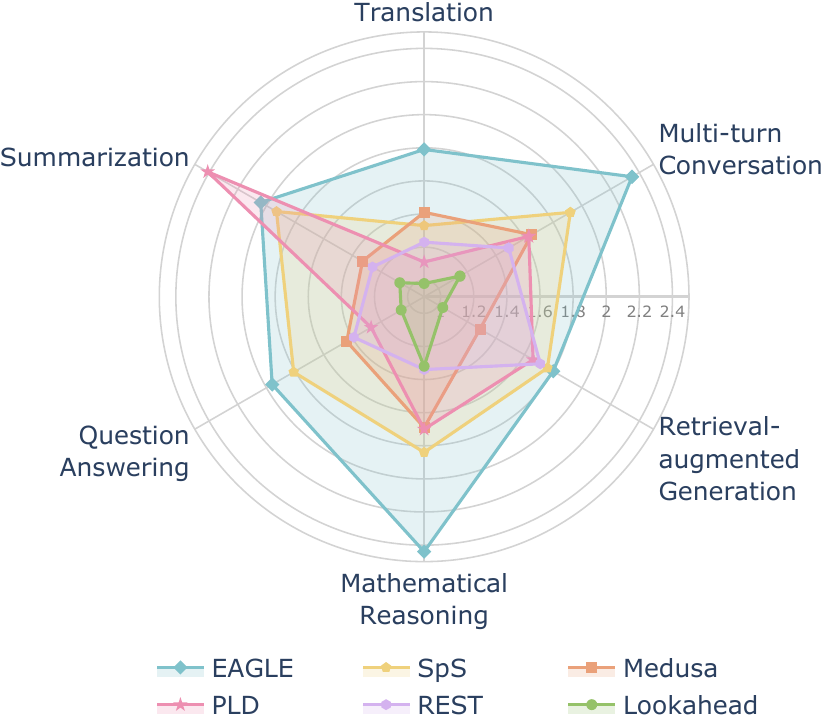}
    \caption{Speedup comparison of various Speculative Decoding methods on Spec-Bench with greedy settings ($T=0$). Evaluations were conducted on Vicuna-7B with a batch size of 1. We present the mean speedup over 3 runs. The detailed results are shown in Appendix~\ref{appendix:main_details}.}
    \label{fig:radar_3090}
\end{figure}

\subsection{Comparative Evaluation}
Our main evaluations were conducted on Vicuna-7B at FP16 precision using a single consumer-grade 3090 GPU\footnote{For comparative analysis on a more powerful A100 GPU, please refer to Appendix~\ref{appendix:A100_analysis}.}. As depicted in Figure~\ref{fig:radar_3090}, under greedy settings, EAGLE~\cite{Li:2024eagle} achieves the highest speedup ratio (1.8$\times$$\sim$2.4$\times$) over autoregressive decoding across most subtasks, especially in mathematical reasoning (with a $\sim$2.4$\times$ speedup). 
EAGLE's success is mainly due to two factors: 1) it reuses the KV cache of LLMs to predict drafted tokens, substantially reducing the drafting computational overhead; 2) compared with Medusa~\cite{medusa}, EAGLE drafts in an autoregressive way, providing more stable and accurate speculation results.
PLD~\cite{saxena2023pld} excels in subtasks with high similarities between input and output, such as summarization (with a $\sim$2.4$\times$ speedup). However, its performance diminishes in other subtasks like translation and question answering, with speedup ratios falling between 1.1$\times$$\sim$1.3$\times$.

\begin{figure}[t]
    \centering
    \includegraphics[width=0.95\columnwidth]{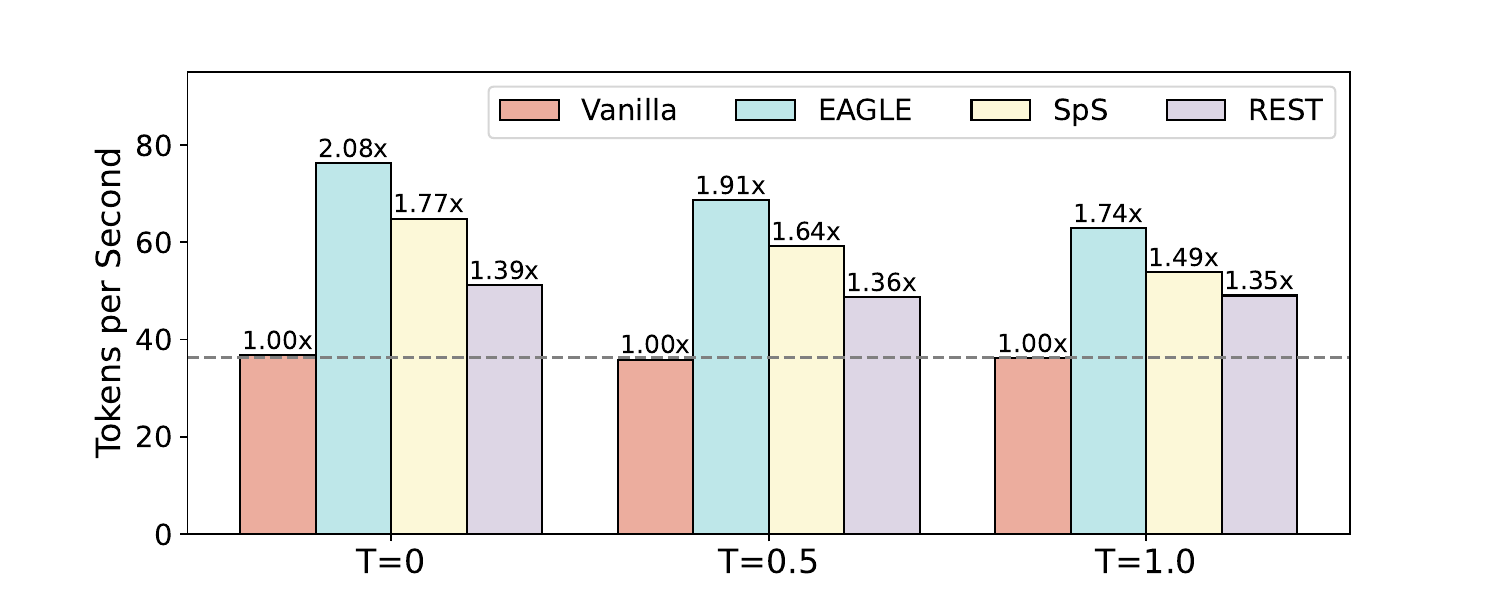}
    \caption{Speedup comparison of various methods on Spec-Bench at different temperatures. The speedup effect diminishes as the sampling temperature increases.}
    \label{fig:sampling}
\end{figure}

We also compare the speedups of Speculative Decoding methods at different sampling temperatures. As illustrated in Figure~\ref{fig:sampling}, EAGLE consistently outperforms other methods across various settings, achieving a speedup ratio ranging from 1.7$\times$ to 2.1$\times$. Besides, it is observed that the acceleration effect of all methods decreases with an increase in sampling temperature. This is attributed to the increased computational complexity of the speculative sampling criterion at higher temperatures, as revealed in prior research~\cite{gante2023assisted, Spector:2023stagedspec}.

\section{Challenges and Future Directions}
\label{sec:directions}

\paragraph{How to trade off speculation accuracy and drafting efficiency?}
As discussed in Sections~\ref{sec:drafting}, scaling up the drafter can effectively enhance speculation accuracy, yet it largely reduces the drafting efficiency and even the overall speedup. Therefore, it is essential to strike a balance between speculation accuracy and drafting latency. Among existing strategies, behavior alignment is a promising approach to address this issue, as it improves speculation accuracy without increasing latency. However, despite recent advancements~\cite{Miao:2023specinfer, Zhou:2023distillspec, Liu:2023onlinespec}, there is still considerable room for improvement to align the drafter with the target LLM. For example, given that the drafted tokens after the bifurcation position are all discarded, one potential direction could involve encouraging the drafter to prioritize the generation quality of early-position tokens. Beyond alignment, other factors such as the quality of drafting~\cite{Fu:2023lookahead} and the determination of speculation length~\cite{Su:2023batching} also influence speculation accuracy and merit further exploration.

\paragraph{How to apply Speculative Decoding in batched inference scenarios?}
Currently, only a few Speculative Decoding implementations have supported batched inference, such as EAGLE\footnote{\url{https://github.com/SafeAILab/EAGLE}} and SpS\footnote{\url{https://github.com/lucidrains/speculative-decoding}}. However, batched inference is a crucial technique for efficiently managing user inputs in LLM real-time services. The primary challenges in batched Speculative Decoding lie in two aspects: (1) Each decoded sentence in Speculative Decoding varies in decoding steps due to different speculation accuracy. Thus, the inference latency of a batch depends on the slowest sample in the batch; (2) The extra computational complexity introduced by Speculative Decoding, especially in sampling settings, increases with larger batch sizes. How to maintain a promising speedup of Speculative Decoding in batched inference, and combine it with advanced techniques such as continuous batching~\cite{Yu:2022continuousbatching}, warrants further investigation.

\paragraph{How to integrate Speculative Decoding with other leading techniques?}
As a general decoding paradigm, Speculative Decoding has already demonstrated its potential in conjunction with other advanced techniques~\cite{yang:2023llma, Zhang2023RaLMSpec, Li:2023contrastive}. For instance, \citet{Yuan:2023SCD} combined Speculative Decoding with Contrastive Decoding~\cite{Li:2023contrastive}, which not only speeds up the inference but also substantially improves the generation quality. In addition to the acceleration of text-only LLMs, applying Speculative Decoding in multimodal inference, such as image synthesis, text-to-speech synthesis, and video generation, is also an intriguing and valuable direction for future research. Another promising research direction is to integrate Speculative Decoding with other efficient methods such as vLLM~\cite{Kwon:2023vLLM}, Non-Auregressive Generation~\cite{Du:2021oaxe, Du:2022ngramxe} and Flash-Attention~\cite{Dao:2022flashattention, Dao:2023flashattention2}, further boosting the inference efficiency of LLM services.
\section{Conclusion}
\label{sec:conclusion}
This paper presents a comprehensive survey of Speculative Decoding, including the evolution of this promising paradigm, its formal definition and formulation, a systematic categorization of existing methods, and an in-depth review of leading techniques. Moreover, we introduce Spec-Bench, an extensive evaluation benchmark for Speculative Decoding methods, and present a comparative evaluation of prominent methods. To our knowledge, this is the first survey dedicated to Speculative Decoding. Our aim for this paper is to clarify the current research landscape and provide insights into future research directions.

\section*{Limitations}
\label{subsec:limitation} 

This paper provides a thorough examination and categorization of current methodologies and emerging trends in Speculative Decoding. We have also conducted a comparative analysis of leading open-source methods to offer researchers deeper insights into the advantages and limitations of different models. Beyond Speculative Decoding, we acknowledge additional efficient NLP strategies such as vLLM~\cite{Kwon:2023vLLM} and continuous batching~\cite{Yu:2022continuousbatching}. In the future, we intend to expand the discussion to encompass the integration of Speculative Decoding with these advanced techniques. Moreover, due to the absence of an available implementation of batched Speculative Decoding, our evaluations could not cover this aspect. We plan to undertake subsequent experiments to assess the speedup of Speculative Decoding methods across various batch sizes.

\section*{Ethics Statement}
\label{subsec:ethics} 
The datasets used in our experiment are publicly released and labeled through interaction with humans in English. In this process, user privacy is protected, and no personal information is contained in the dataset. The scientific artifacts that we used are available for research with permissive licenses. And the use of these artifacts in this paper is consistent with their intended use. Therefore, we believe that our research work meets the ethics of ACL. 

\section*{Acknowledgements}
We thank all anonymous reviewers for their valuable comments during the review process. This work is partially supported by Research Grants Council of Hong Kong (15207122 and 15213323).

\bibliography{anthology,custom}

\clearpage

\appendix

\section*{Appendix}
\section{Applications}
\label{sec:applications}
In addition to serving as a general paradigm, recent work has revealed that some variants of Speculative Decoding demonstrate extraordinary effectiveness in specific tasks. Furthermore, other research has applied this paradigm to address latency issues unique to certain application scenarios, achieving inference acceleration. Below, we will provide a detailed introduction to these promising works.

Recent studies have highlighted Speculative Decoding is particularly well suited for tasks where model inputs and outputs are highly similar~\cite{Sun:2020SAD, ge2022lossless, yang:2023llma}, such as Grammatical Error Correction~\cite{Wang:2021gec, Bryant:2023gec} and Retrieval-augmented Generation~\cite{Cai:2022rag}. These methods introduced a specialized form of Speculative Decoding, where the initial user input or the retrieved context is directly employed as drafts. For instance, SAD~\cite{Sun:2020SAD}, an early attempt at Speculative Decoding on Grammatical Error Correction, utilized the input sentence with grammatical errors as a draft and leveraged the LLM to verify the whole sentence in parallel, achieving a $9\times$$\sim$$12\times$ speedup. Similarly, LLMA~\cite{yang:2023llma} selected text spans from the reference as drafts, demonstrating a $2\times$$\sim$$3\times$ speedup across various practical application scenarios including Retrieval-augmented Generation, Cache-assisted Generation, and Multi-turn Conversations.

Beyond these works, RaLMSpec~\cite{Zhang2023RaLMSpec} adopted Speculative Decoding to accelerate retrieval-augmented language models (RaLMs). It pointed out that the main latency bottleneck of iterative RaLMs is the frequent retrieval from a vast knowledge base. To accelerate inference, this method proposed to maintain a local cache for speculative retrieval, achieving around $2\times$ speedup with identical model outputs. LLMCad~\cite{Xu:2023LLMCad} applied Speculative Decoding to on-device LLM inference. Concretely, it proposed to generate drafts with a smaller real-time LM that can be hosted in device memory, and only utilize the target LLM for parallel verification. This approach effectively reduces repetitive releasing and loading of model weights, achieving a $9.3\times$ speedup compared to existing inference engines.
\section{Experimental Details}
\label{appendix:experimental_details}

\subsection{Details of Spec-Bench}
\label{appendix:spec_bench}
To assess the acceleration performance of Speculative Decoding methods in various scenarios, we developed Spec-Bench, a comprehensive benchmark encompassing six distinct tasks. Spec-Bench integrates MT-bench~\cite{zheng2023mtbench}, a multi-turn conversation benchmark previously adopted in research~\cite{medusa, Li:2024eagle}, to provide a basis for comparison with earlier studies. Additionally, it includes two input-guided tasks: summarization and retrieval-augmented generation (RAG), both of which exhibit a significant overlap between the input prompts and the target outputs. We selected CNN/Daily Mail~\cite{Nallapati:2016cnndm} and Natural Questions~\cite{kwiatkowski2019:naturalquestions} as the dataset for these two tasks, respectively. Specifically, in the RAG subtask, the top-5 documents retrieved from DPR~\cite{Karpukhin:2020dpr} were concatenated with each question to construct the input prompt.

Moreover, Spec-Bench incorporates three further subtasks -- translation, question answering, and mathematical reasoning -- to provide a thorough evaluation of Speculative Decoding's speedup capabilities in diverse contexts. We utilized WMT14 DE-EN, Natural Questions, and GSM8K~\cite{Cobbe:2021gsm8k} as the primary datasets for these tasks, respectively. We randomly selected 80 instances from each subtask's test set for evaluation. The detailed composition is summarized in Table~\ref{tab:spec_bench}.

\begin{table}[htbp]
\centering
\small
\resizebox{\linewidth}{!}{
\begin{tabular}{ll|c}
\toprule
\bf Subtask &\bf Dataset &\bf \#Samples\\\midrule
Multi-turn Conversation &MT-bench &80\\
Retrieval-aug. Generation &Natural Questions &80 \\
Summarization &CNN/Daily Mail &80 \\
Translation &WMT14 DE-EN &80 \\
Question Answering &Natural Questions &80 \\
Mathematical Reasoning &GSM8K &80 \\\midrule
Overall &- &480 \\
\bottomrule
\end{tabular}}
\caption{Detailed Composition of Spec-Bench. Spec-Bench includes 6 distinct subtasks to encompass diverse application scenarios.}
\label{tab:spec_bench}
\end{table}
\begin{table*}[t!]
\small 
\centering
\resizebox{\linewidth}{!}{
\begin{tabular}{@{}ll|cccccc|cc@{}}
\toprule
\multicolumn{2}{l|}{\bf Models} &\begin{tabular}[c]{@{}c@{}}\bf Multi-turn\\\bf Conversation\end{tabular} &\bf Translation & \bf Summarization & \begin{tabular}[c]{@{}c@{}}\bf Question\\\bf Answering\end{tabular} & \begin{tabular}[c]{@{}c@{}}\bf Mathematical\\\bf Reasoning\end{tabular} & \begin{tabular}[c]{@{}c@{}}\bf Retrieval-aug.\\\bf Generation\end{tabular} & \bf \#tokens/s & \bf Avg. \\
\midrule
\multirow{7}{*}{\rotatebox{90}{$T=0$}} 
&Autoregressive Decoding &1.00$\times$\cpm{0.00} &1.00$\times$\cpm{0.00}	&1.00$\times$\cpm{0.00} &1.00$\times$\cpm{0.00} &1.00$\times$\cpm{0.00} &1.00$\times$\cpm{0.00} &36.74\cpm{0.31} &1.00$\times$ \\
&Lookahead~\cite{Fu:2023lookahead} &1.15$\times$\cpm{0.01} &0.98$\times$\cpm{0.02} &1.07$\times$\cpm{0.02} &1.06$\times$\cpm{0.02} &1.32$\times$\cpm{0.02} &1.03$\times$\cpm{0.02}	&40.64\cpm{0.26} &1.11$\times$\\ 
&REST~\cite{He:2023REST} &1.49$\times$\cpm{0.02} &1.23$\times$\cpm{0.04} &1.26$\times$\cpm{0.03}	&1.39$\times$\cpm{0.04} &1.34$\times$\cpm{0.03} &1.71$\times$\cpm{0.05} &51.12\cpm{0.78} &1.39$\times$\\ 
&PLD~\cite{saxena2023pld} &1.63$\times$\cpm{0.02} &1.11$\times$\cpm{0.02} &\bf 2.41$\times$\cpm{0.04}	&1.27$\times$\cpm{0.03} &1.70$\times$\cpm{0.03} &1.66$\times$\cpm{0.04} &59.42\cpm{0.55} &1.62$\times$\\ 
&SpS~\cite{Leviathan:2023specdec} &1.92$\times$\cpm{0.04} &1.33$\times$\cpm{0.02} &1.93$\times$\cpm{0.01}	&1.81$\times$\cpm{0.04} &1.84$\times$\cpm{0.00} &1.76$\times$\cpm{0.01} &64.85\cpm{0.70} &1.77$\times$\\ 
&Medusa~\cite{medusa} &1.65$\times$\cpm{0.03} &1.41$\times$\cpm{0.02} &1.33$\times$\cpm{0.01}	&1.44$\times$\cpm{0.03} &1.69$\times$\cpm{0.01} &1.29$\times$\cpm{0.02} &54.30\cpm{0.34} &1.48$\times$\\ 
&EAGLE~\cite{Li:2024eagle}  &\bf 2.35$\times$\cpm{0.03} &\bf 1.79$\times$\cpm{0.03} &2.04$\times$\cpm{0.02} &\bf 1.96$\times$\cpm{0.03} &\bf 2.44$\times$\cpm{0.02} &\bf 1.80$\times$\cpm{0.03} &\bf 76.30\cpm{0.36} &\bf 2.08$\times$\\\midrule
\multirow{4}{*}{\rotatebox{90}{$T=1$}} 
&Autoregressive Decoding &1.00$\times$\cpm{0.00} &1.00$\times$\cpm{0.00}	&1.00$\times$\cpm{0.00} &1.00$\times$\cpm{0.00} &1.00$\times$\cpm{0.00} &1.00$\times$\cpm{0.00} &36.24\cpm{0.43} &1.00$\times$ \\
&REST~\cite{He:2023REST} &1.43$\times$\cpm{0.01} &1.19$\times$\cpm{0.02} &1.24$\times$\cpm{0.00}	&1.36$\times$\cpm{0.02} &1.34$\times$\cpm{0.02} &1.61$\times$\cpm{0.02} &49.04\cpm{0.30} &1.35$\times$\\ 
&SpS~\cite{Leviathan:2023specdec} &1.55$\times$\cpm{0.01} &1.20$\times$\cpm{0.01} &1.57$\times$\cpm{0.01}	&1.54$\times$\cpm{0.03} &1.56$\times$\cpm{0.03} &1.52$\times$\cpm{0.02} &53.94\cpm{0.43} &1.49$\times$\\ 
&EAGLE~\cite{Li:2024eagle}  &\bf 1.79$\times$\cpm{0.02} &\bf 1.61$\times$\cpm{0.03} &\bf 1.74$\times$\cpm{0.03} &\bf 1.66$\times$\cpm{0.04} &\bf 1.95$\times$\cpm{0.06} &\bf 1.63$\times$\cpm{0.03} &\bf 62.88\cpm{0.54} &\bf 1.74$\times$\\
\bottomrule
\end{tabular}}
\caption{Speedup comparison of various Speculative Decoding methods on Spec-Bench. The results were obtained using \texttt{Vicuna-7B-v1.3} at FP16 precision. Evaluations were conducted on a single NVIDIA 3090 GPU with a batch size of 1. We report the mean speedup ratio over 3 different runs. We show the best results in \textbf{boldface}.}
\label{tab:3090_detail}
\end{table*}

\subsection{Implementation Details}
We have selected six representative Speculative Decoding methods for our comparative analysis on Spec-Bench. These methods are open-source and free of bugs. Specifically, \textbf{SpS}~\cite{Chen:2023specsampling} stands as the pioneering work in this field, utilizing a smaller LM from the same model series as the drafter to accelerate LLM inference. \textbf{Medusa}~\cite{medusa} and \textbf{EAGLE}~\cite{Li:2024eagle} integrate additional lightweight heads into the target LLM to facilitate efficient drafting. \textbf{Lookahead}~\cite{Fu:2023lookahead} introduces multiple special tokens to the end of the input prompt for parallel drafting and transforms the drafts into n-gram candidates. \textbf{PLD}~\cite{saxena2023pld} is the code implementation\footnote{\url{https://github.com/apoorvumang/prompt-lookup-decoding}} of LLMA~\cite{yang:2023llma}, which selects text spans from the input as drafts. \textbf{REST}~\cite{He:2023REST} retrieves relevant drafts from text corpora based on the input prompt.

We conducted our experimental evaluations using the \texttt{Vicuna-v1.3} model series~\cite{zheng2023mtbench}. For SpS, we employed the Huggingface implementation\footnote{\url{https://huggingface.co/blog/assisted-generation}} and utilized the \texttt{vicuna-68m-v1.3} model provided by \citet{Yang:2024mcsd} as the drafter. We followed the default parameters of Lookahead\footnote{\url{https://github.com/hao-ai-lab/LookaheadDecoding}} and PLD for our evaluations. The main experiments were conducted using Pytorch 2.0.1 with a single consumer-grade NVIDIA GeForce RTX 3090 GPU (24GB) of 12 CPU cores under CUDA 11.8. Further analysis was performed on a more powerful NVIDIA A100 GPU (80GB) of 64 CPU cores under CUDA 11.4.
\section{Details of Main Experimental Results}
\label{appendix:main_details}
The detailed results of our main analysis are shown in Table~\ref{tab:3090_detail}, including the experimental settings of greedy decoding ($T=0$) and speculative sampling ($T=1$). The findings indicate that EAGLE~\cite{Li:2024eagle} excels across various Spec-Bench subtasks, achieving an overall speedup ranging from 1.6$\times$ to 2.4$\times$. PLD~\cite{saxena2023pld} shows notable efficiency in scenarios where the input and output have a significant overlap. For instance, the speedup ratio of PLD increases from 1.27$\times$ in the question answering subtask to 1.66$\times$ in the retrieval-augmented generation subtask, highlighting its effectiveness when the input includes relevant documents. Notably, most methods achieve a suboptimal speedup on the translation subtask. We suspect that it is due to the potential lack of multilingual data in the pretraining corpora.
\begin{figure}[t]
\centering
    \includegraphics[width=0.95\columnwidth]{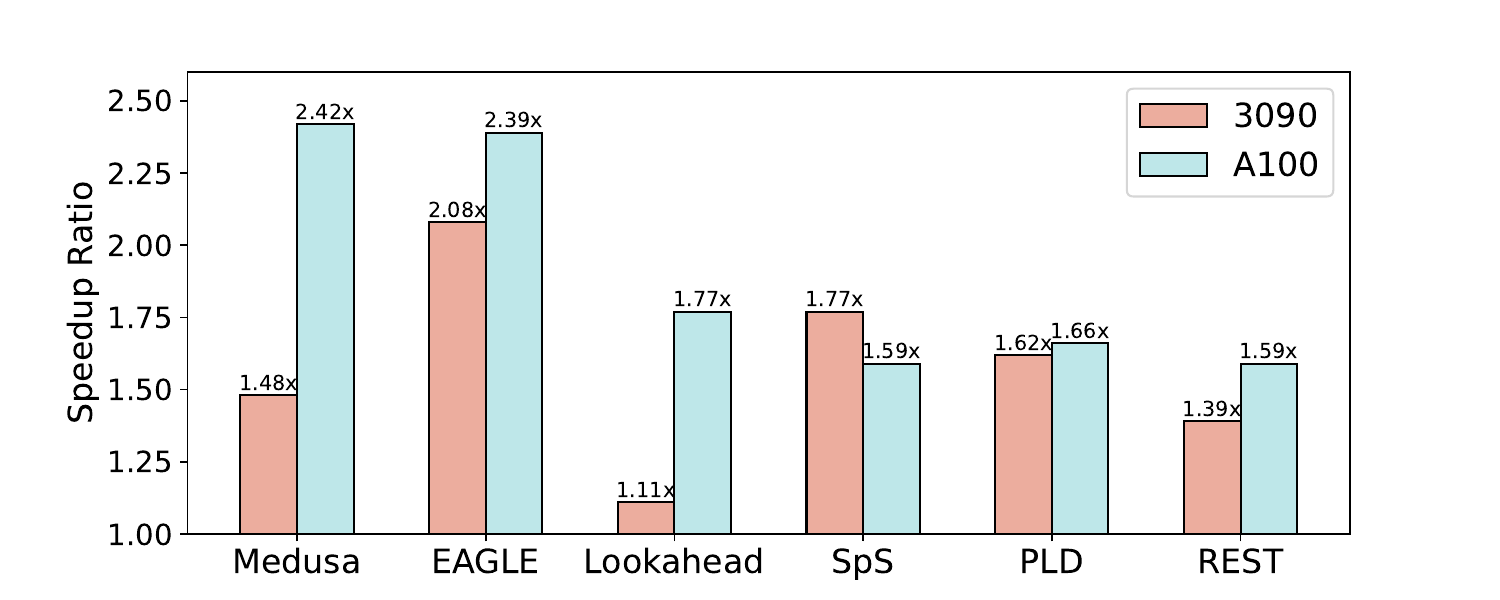}
    \caption{Speedup comparison of various methods on Spec-Bench with different computational devices.}
    \label{fig:device}
\end{figure}

\begin{table*}[t!]
\small 
\centering
\resizebox{\linewidth}{!}{
\begin{tabular}{@{}ll|cccccc|cc@{}}
\toprule
\multicolumn{2}{l|}{\bf Models} &\begin{tabular}[c]{@{}c@{}}\bf Multi-turn\\\bf Conversation\end{tabular} &\bf Translation & \bf Summarization & \begin{tabular}[c]{@{}c@{}}\bf Question\\\bf Answering\end{tabular} & \begin{tabular}[c]{@{}c@{}}\bf Mathematical\\\bf Reasoning\end{tabular} & \begin{tabular}[c]{@{}c@{}}\bf Retrieval-aug.\\\bf Generation\end{tabular} & \bf \#tokens/s & \bf Avg. \\
\midrule
\multirow{7}{*}{\rotatebox{90}{\texttt{Vicuna-7B}}} 
&Autoregressive Decoding &1.00$\times$\cpm{0.00} &1.00$\times$\cpm{0.00}	&1.00$\times$\cpm{0.00} &1.00$\times$\cpm{0.00} &1.00$\times$\cpm{0.00} &1.00$\times$\cpm{0.00} &40.24\cpm{0.30} &1.00$\times$ \\
&Lookahead~\cite{Fu:2023lookahead} &1.95$\times$\cpm{0.01} &1.61$\times$\cpm{0.05} &1.63$\times$\cpm{0.02} &1.73$\times$\cpm{0.04} &2.16$\times$\cpm{0.04} &1.50$\times$\cpm{0.00}	&71.20\cpm{1.30} &1.77$\times$\\ 
&REST~\cite{He:2023REST} &1.72$\times$\cpm{0.06} &1.38$\times$\cpm{0.05} &1.46$\times$\cpm{0.04}	&1.80$\times$\cpm{0.04} &1.31$\times$\cpm{0.03} &1.87$\times$\cpm{0.06} &63.81\cpm{1.00} &1.59$\times$\\ 
&PLD~\cite{saxena2023pld} &1.67$\times$\cpm{0.03} &1.06$\times$\cpm{0.03} &\bf 2.59$\times$\cpm{0.06}	&1.16$\times$\cpm{0.03} &1.63$\times$\cpm{0.03} &1.83$\times$\cpm{0.02} &66.61\cpm{1.15} &1.66$\times$\\ 
&SpS~\cite{Leviathan:2023specdec} &1.78$\times$\cpm{0.03} &1.19$\times$\cpm{0.02} &1.78$\times$\cpm{0.03}	&1.58$\times$\cpm{0.03} &1.54$\times$\cpm{0.02} &1.69$\times$\cpm{0.02} &64.07\cpm{0.41} &1.59$\times$\\ 
&Medusa~\cite{medusa} &\bf 2.79$\times$\cpm{0.07} &\bf 2.36$\times$\cpm{0.07} &2.14$\times$\cpm{0.04}	&\bf 2.36$\times$\cpm{0.08} &2.77$\times$\cpm{0.08} &2.05$\times$\cpm{0.01} &\bf 97.27\cpm{2.04} &\bf 2.42$\times$\\ 
&EAGLE~\cite{Li:2024eagle}  &2.75$\times$\cpm{0.05} &2.08$\times$\cpm{0.05} &2.32$\times$\cpm{0.05} &2.23$\times$\cpm{0.03} &\bf 2.79$\times$\cpm{0.04} &\bf 2.15$\times$\cpm{0.01} &96.23\cpm{1.15} &2.39$\times$\\
\midrule
\multirow{7}{*}{\rotatebox{90}{\texttt{Vicuna-13B}}} 
&Autoregressive Decoding &1.00$\times$\cpm{0.00} &1.00$\times$\cpm{0.00} &1.00$\times$\cpm{0.00} &1.00$\times$\cpm{0.00} &1.00$\times$\cpm{0.00} &1.00$\times$\cpm{0.00} &31.38\cpm{0.22} &1.00$\times$ \\
&Lookahead~\cite{Fu:2023lookahead} &1.57$\times$\cpm{0.01} &1.34$\times$\cpm{0.01} &1.39$\times$\cpm{0.00} &1.40$\times$\cpm{0.01} &1.82$\times$\cpm{0.02} &1.32$\times$\cpm{0.01}	&46.42\cpm{0.12} &1.48$\times$\\ 
&REST~\cite{He:2023REST} &1.68$\times$\cpm{0.01} &1.31$\times$\cpm{0.05} &1.51$\times$\cpm{0.01}	&1.67$\times$\cpm{0.02} &1.29$\times$\cpm{0.00} &1.96$\times$\cpm{0.01} &48.89\cpm{0.26} &1.56$\times$\\ 
&PLD~\cite{saxena2023pld} &1.53$\times$\cpm{0.02} &1.08$\times$\cpm{0.01} &2.25$\times$\cpm{0.00}	&1.09$\times$\cpm{0.02} &1.65$\times$\cpm{0.03} &1.72$\times$\cpm{0.00} &48.42\cpm{0.17} &1.54$\times$\\ 
&SpS~\cite{Leviathan:2023specdec} &1.73$\times$\cpm{0.02} &1.25$\times$\cpm{0.02} &1.76$\times$\cpm{0.00}	&1.53$\times$\cpm{0.01} &1.68$\times$\cpm{0.00} &1.73$\times$\cpm{0.00} &50.48\cpm{0.28} &1.61$\times$\\ 
&Medusa~\cite{medusa} &2.39$\times$\cpm{0.02} &2.12$\times$\cpm{0.02} &1.92$\times$\cpm{0.00}	&2.07$\times$\cpm{0.02} &2.49$\times$\cpm{0.02} &1.88$\times$\cpm{0.00} &67.64\cpm{0.07} &2.16$\times$\\ 
&EAGLE~\cite{Li:2024eagle}  &\bf 2.88$\times$\cpm{0.05} &\bf 2.24$\times$\cpm{0.04} &\bf 2.52$\times$\cpm{0.03} &\bf 2.24$\times$\cpm{0.04} &\bf 2.90$\times$\cpm{0.03} &\bf 2.34$\times$\cpm{0.01} &\bf 79.35\cpm{1.18} &\bf 2.53$\times$\\
\midrule
\multirow{7}{*}{\rotatebox{90}{\texttt{Vicuna-33B}}} 
&Autoregressive Decoding &1.00$\times$\cpm{0.00} &1.00$\times$\cpm{0.00}	&1.00$\times$\cpm{0.00} &1.00$\times$\cpm{0.00} &1.00$\times$\cpm{0.00} &1.00$\times$\cpm{0.00} &16.34\cpm{0.01} &1.00$\times$ \\
&Lookahead~\cite{Fu:2023lookahead} &1.46$\times$\cpm{0.00} &1.21$\times$\cpm{0.00} &1.32$\times$\cpm{0.00} &1.29$\times$\cpm{0.00} &1.71$\times$\cpm{0.00} &1.28$\times$\cpm{0.00}	&22.58\cpm{0.08} &1.38$\times$\\ 
&REST~\cite{He:2023REST} &1.71$\times$\cpm{0.01} &1.39$\times$\cpm{0.00} &1.57$\times$\cpm{0.00}	&1.69$\times$\cpm{0.01} &1.34$\times$\cpm{0.01} &1.89$\times$\cpm{0.00} &25.98\cpm{0.07} &1.59$\times$\\ 
&PLD~\cite{saxena2023pld} &1.45$\times$\cpm{0.00} &1.06$\times$\cpm{0.00} &1.98$\times$\cpm{0.00}	&1.07$\times$\cpm{0.00} &1.54$\times$\cpm{0.00} &1.43$\times$\cpm{0.00} &23.07\cpm{0.01} &1.41$\times$\\ 
&SpS~\cite{Leviathan:2023specdec} &1.79$\times$\cpm{0.00} &1.31$\times$\cpm{0.00} &1.80$\times$\cpm{0.00}	&1.57$\times$\cpm{0.00} &1.73$\times$\cpm{0.00} &1.69$\times$\cpm{0.00} &26.89\cpm{0.03} &1.65$\times$\\ 
&Medusa~\cite{medusa} &2.22$\times$\cpm{0.00} &1.95$\times$\cpm{0.00} &1.85$\times$\cpm{0.00}	&1.87$\times$\cpm{0.01} &2.32$\times$\cpm{0.01} &1.84$\times$\cpm{0.00} &32.92\cpm{0.06} &2.01$\times$\\ 
&EAGLE~\cite{Li:2024eagle}  &\bf 2.81$\times$\cpm{0.00} &\bf 2.14$\times$\cpm{0.00} &\bf 2.53$\times$\cpm{0.00} &\bf 2.19$\times$\cpm{0.00} &\bf 3.01$\times$\cpm{0.00} &\bf 2.31$\times$\cpm{0.00} &\bf 40.91\cpm{0.03} &\bf 2.50$\times$\\
\bottomrule
\end{tabular}}
\caption{Speedup comparison of Speculative Decoding methods across various model scales on Spec-Bench. The results were obtained using \texttt{Vicuna-v1.3} at FP16 precision with greedy settings ($T=0$). Evaluations were conducted on a single NVIDIA A100 GPU with a batch size of 1. We report the mean speedup over 3 different runs.}
\label{tab:A100_scale}
\end{table*}

\section{Further Analysis on A100}
\label{appendix:A100_analysis}
This section presents a comprehensive analysis of leading Speculative Decoding methods on Spec-Bench, utilizing a single NVIDIA A100 GPU. The discussion delves into the influence of computational hardware, model scale, and computational precision on the performance of Speculative Decoding. All experiments were performed on \textit{the same device} and \textit{environment} to ensure fair comparison.

\begin{figure}[t]
\centering
    \includegraphics[width=0.95\columnwidth]{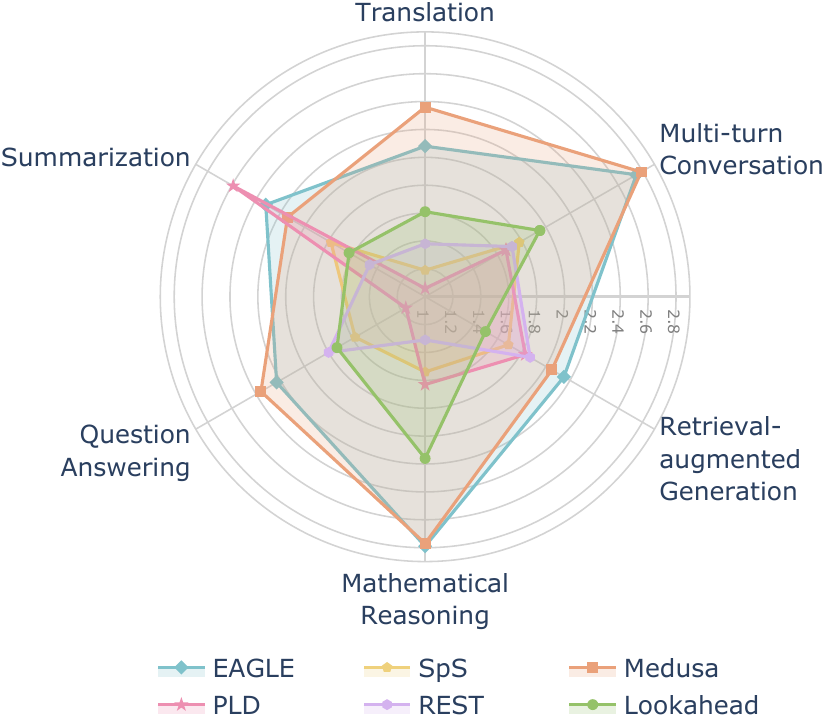}
    \caption{Speedup comparison of various Speculative Decoding methods on a single A100 GPU with greedy settings ($T=0$). Evaluations were conducted on Spec-Bench using Vicuna-7B at FP16 precision.}
    \label{fig:radar_A100}
\end{figure}

\subsection{Computational Devices}
We first discuss the impact of evolving computational devices on Speculative Decoding. As depicted in Figure~\ref{fig:device}, the acceleration effect of most Speculative Decoding methods is notably enhanced when employed on high-performance GPUs, such as NVIDIA A100s. This enhancement is primarily due to the increased availability of idle computational resources on more advanced computational devices, which Speculative Decoding can leverage to accelerate inference processes. Among the methods evaluated, Medusa~\cite{medusa} and Lookahead~\cite{Fu:2023lookahead} demonstrate the most significant improvements. Specifically, the speedup ratio for Medusa escalates from 1.48$\times$ to 2.42$\times$, and for Lookahead, it rises from 1.11$\times$ to 1.77$\times$. This finding underscores that Speculative Decoding methods will benefit more from evolving computational hardware, such as H100 GPUs.

We illustrate the comparison of various Speculative Decoding methods evaluated with a single A100 GPU in Figure~\ref{fig:radar_A100}. The detailed experimental results are shown in Table~\ref{tab:A100_scale}. The results indicate that Medusa~\cite{medusa} and EAGLE~\cite{Li:2024eagle} excel in this experimental setting, achieving an overall speedup of 2.4$\times$. These two methods perform particularly well on the multi-turn conversation and mathematical reasoning subtasks, with a $\sim$2.8$\times$ speedup.

\begin{figure}[t]
\centering
    \includegraphics[width=0.95\columnwidth]{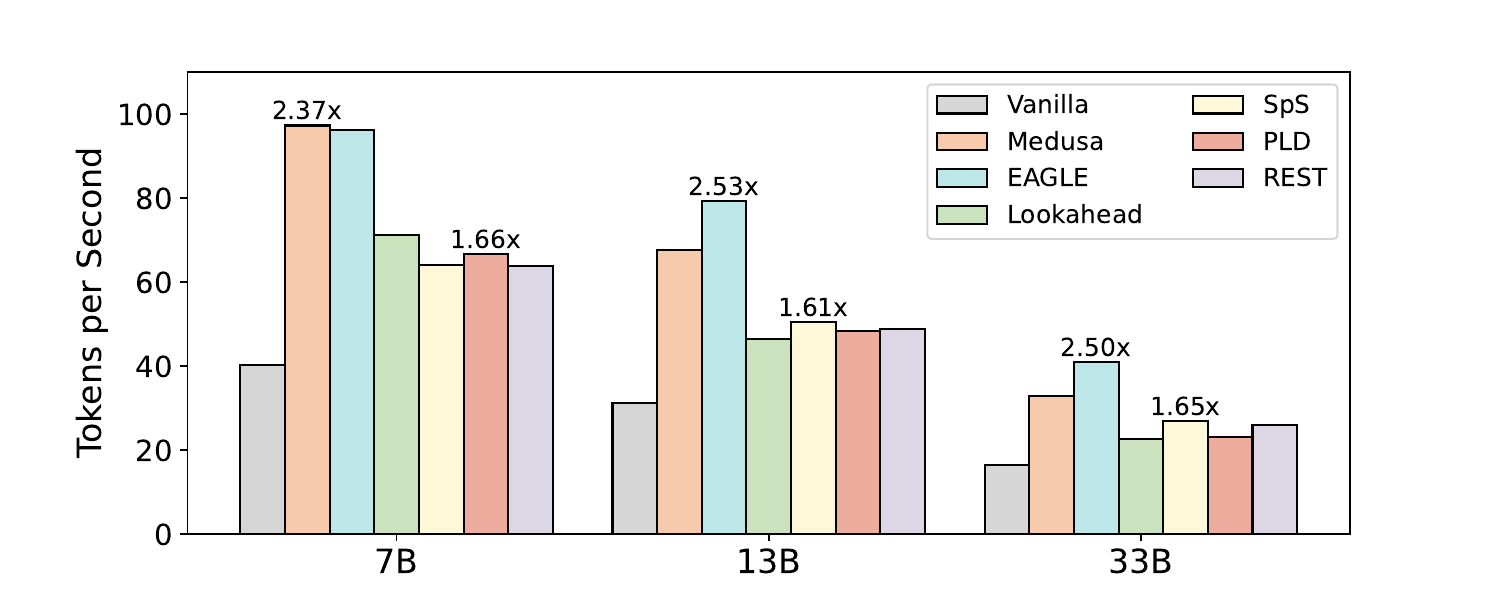}
    \caption{Speedup comparison of various methods on Spec-Bench at different model scales.}
    \label{fig:model_scale}
\end{figure}

\subsection{Model Scale}
We present the speedup comparison of Speculative Decoding methods across various model scales in Figure~\ref{fig:model_scale}. The detailed experimental results are shown in Table~\ref{tab:A100_scale}. Among all the evaluated methods, EAGLE~\cite{Li:2024eagle} maintains a high speedup ratio over autoregressive decoding across all model scales, achieving a speedup ratio ranging from 2.4$\times$ to 2.5$\times$. While Medusa~\cite{medusa} demonstrates superior acceleration performance on Vicuna-7B, its speedup ratio degrades from 2.4$\times$  to 2.0$\times$ as the model scale increases.

\subsection{Computational Precision}
\label{appdix:precision}
It is noteworthy that most Speculative Decoding approaches are predominantly evaluated using FP16 precision~\cite{Fu:2023lookahead, medusa, Li:2024eagle, He:2023REST}. However, it is critical to underscore that the outputs generated by Speculative Decoding in FP16 precision may not consistently align with those derived from autoregressive decoding. This divergence stems from the accumulation of floating-point errors inherent in FP16 computations, which can result in discrepancies between the outputs of the two decoding methods, particularly in the context of longer sequences. In FP32 precision, the outputs of Speculative Decoding are guaranteed to be exactly the same as autoregressive decoding.

\begin{figure}[t]
\centering
    \includegraphics[width=0.95\columnwidth]{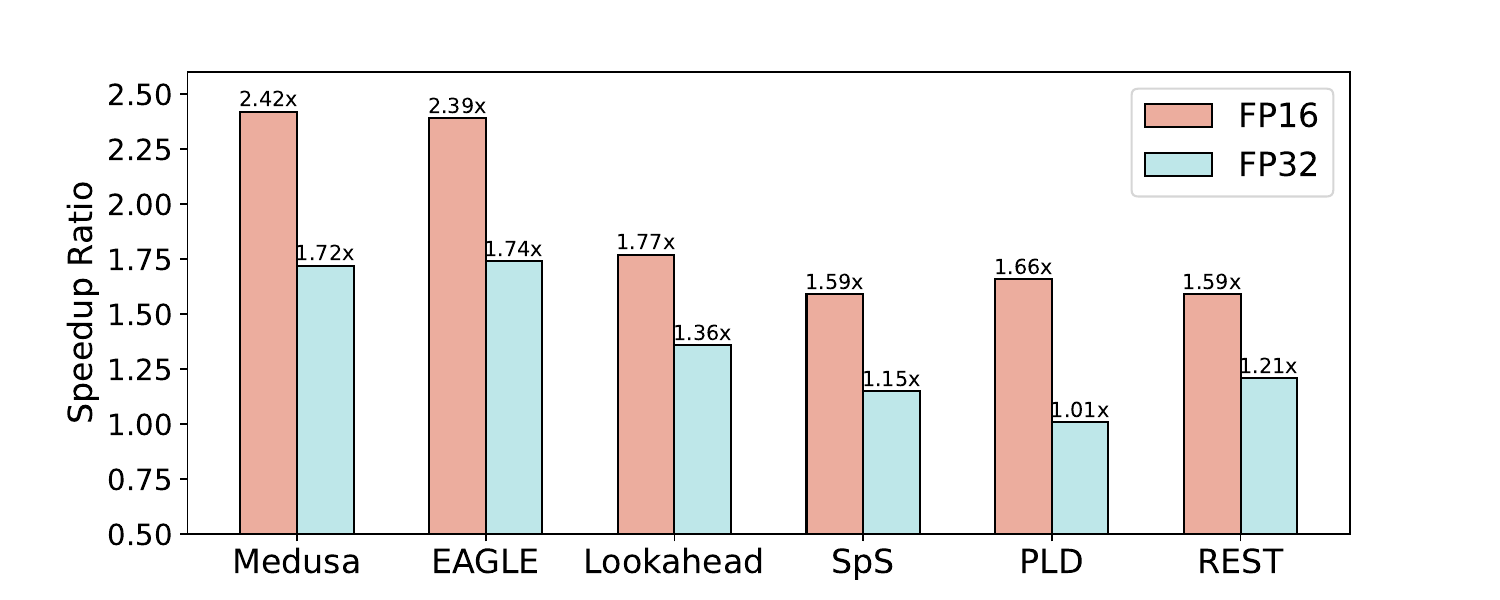}
    \caption{Speedup comparison of various methods on Spec-Bench with different computational precision.}
    \label{fig:precision}
\end{figure}

We compare the speedup performance of Speculative Decoding methods with FP16/FP32 precision in Figure~\ref{fig:precision}. The experimental results reveal a noticeable reduction in speedup for all methods under FP32 precision. Specifically, PLD~\cite{saxena2023pld} achieves merely 1.01$\times$ speedup in FP32 precision, and the acceleration effect of EAGLE~\cite{Li:2024eagle} also diminishes, with its speedup falling from 2.39$\times$ to 1.74$\times$. To furnish the research community with a comprehensive understanding of the acceleration impact, we advocate for future studies to report speedup metrics across both precision settings.

\end{document}